\documentclass{article}
\usepackage[final]{corl_2026}
\hypersetup{pdftitle={CuraEval: Evaluating Demonstration Curation on Mixed-Skill Robot Data},
  pdfauthor={Jiuyi Xu, Meida Chen, Yang Ye, Yangming Shi},
  pdfsubject={Oops, I Erred Workshop at CoRL 2026}}
\usepackage{graphicx,booktabs,amsmath,amssymb,placeins,array}

\newcommand{\ph}{p_{\mathrm H}}
\newcommand{\hg}{HG-WT}
\newcommand{\sig}{SigEnt}
\title{CuraEval: Evaluating Demonstration Curation\\
on Mixed-Skill Robot Data}
\author{
  Jiuyi Xu$^{1}$, Meida Chen$^{2}$, Yang Ye$^{3}$, Yangming Shi$^{1}$\\
  \small $^{1}$Colorado School of Mines \quad $^{2}$University of Southern California -- Institute for Creative Technologies\\
  \small $^{3}$Northeastern University
}

\begin{document}
\maketitle
\begin{abstract}
Robot imitation learning uses demonstrations from operators of varying skill, but selecting training data remains challenging.
Existing methods score demonstrations using smoothness, diversity, or estimated influence on policy performance, but differences in selection and evaluation protocols make it difficult to determine whether they outperform simple baselines.
We introduce \textbf{CuraEval}, a framework combining mathematical analysis of demonstration scores with controlled comparisons against fixed size-matched random subsets and full-data training on mixed-skill robomimic data using Diffusion Policy and BC-RNN.
We show that the evaluated PSD score depends only on episode length and squared position magnitudes, and produces nearly the same ranking as episode length on \textit{square} ($\rho=0.999$).
With Diffusion Policy, selecting the shortest 50\% of demonstrations improves \textit{square} success from 70.7\% under random selection to 76.5\%.
Compared with random selection, our adapted geometric-entropy LOO selector reduces success at 25\% retention from 94.7\% to 74.0\% on \textit{can} and achieves 41.7\% on \textit{square}, below the mean success rates of all three independently sampled random subsets (59.8--62.5\%). 
On the same 3 \textit{square} Diffusion Policy training seeds, length-based selection at 75\% retention improves success over full-data training from 76.7\% to 86.7\% under the original protocol, but lowers it from 80.5\% to 79.5\% after checkpoint reporting, rollout count, and evaluation seed change jointly. 
No evaluated curated condition significantly outperforms full-data training after Holm correction, although confidence intervals still permit useful gains.
Code is available at \url{https://github.com/jiuyixu25/CuraEval}.
\end{abstract}

\keywords{demonstration curation, mixed-skill data, imitation learning}
\section{Introduction}
Imitation learning enables robots to learn manipulation skills from human demonstrations, but performance depends on the quality of the training data~\citep{robomimic,belkhale}.
Even successful demonstrations vary in speed, consistency, and operator skill.
Selecting smooth or efficient trajectories may remove useful coverage, making it difficult to identify the demonstrations that best support learning.

Existing methods evaluate demonstrations through trajectory statistics~\citep{psd,rinse}, diversity~\citep{hg,faktual}, state--action mutual information~\citep{deminf}, or estimated influence on policy performance~\citep{cupid}.
These criteria already inform robot-data pipelines: JoyAI-RA 0.5~\citep{joyaira} uses spectral and jerk-based metrics, while WARP-RM~\citep{warprm} groups T-shirt-folding demonstrations into quality tiers using episode duration.
However, differences in scoring objectives, selection procedures, and evaluation protocols complicate comparisons.
Although several methods include random-selection controls~\citep{rinse,faktual,cupid,deminf}, their behavior under common conditions requires further investigation.
Building on studies of injected demonstration defects~\citep{bedi1,bedi2,bedi3}, we examine natural variation among human operators.

We introduce \textbf{CuraEval}, a framework combining mathematical analysis of demonstration scores with controlled evaluation on state-based, mixed-skill robomimic data~\citep{robomimic}.
We evaluate trajectory scores, explicit adaptations of diversity measures, and our implementation of CUPID~\citep{cupid}, using Diffusion Policy (DP)~\citep{dp} and selected BC-RNN comparisons~\citep{robomimic}.
Fixed training budgets and comparisons with random subsets, simple heuristics, and all data distinguish three goals: selecting better data than random, improving over full-data training, and retaining performance with fewer demonstrations.

Our results show why these distinctions matter.
The evaluated power spectral density (PSD) score closely tracks episode length, and simple heuristics can outperform random selection.
Conversely, our adapted geometric-entropy selector based on leave-one-out (LOO) contributions achieves 41.7\% on \textit{square}, below all three tested random-subset means (59.8--62.5\%).
No curated condition has a Holm-adjusted advantage over full data, while changing the evaluation protocol reverses an initially positive estimate of curation benefit.
Our contributions are:
\begin{itemize}
\item \textbf{Score analysis.} We derive the evaluated PSD score's exact time-domain expression and explain its relationship with episode length.
\item \textbf{Controlled selection evaluation.} We identify beneficial and harmful selections through comparisons with random subsets, simple heuristics, and full data.
\item \textbf{Protocol sensitivity.} We show that evaluation choices can reverse the estimated benefit of curation on the same training seeds.
\end{itemize}

\section{Related Work}
\textbf{Trajectory statistics and diversity.}
PSD-based scoring~\citep{psd} and RINSE's spectral arc length (SAL)~\citep{rinse} use trajectory statistics to assess demonstrations.
The PSD study compares retained-demonstration quality against random removal, but its downstream policy comparison has no random-subset baseline.
Geometric Entropy~\citep{hg} reports a non-monotonic relationship between diversity and policy performance.
FAKTUAL~\citep{faktual} selects subsets using signature-based objectives and notes that low-budget selection can underperform random selection on difficult tasks.
We distinguish these original methods from our explicitly scoped adaptations.

\textbf{Information and influence.}
DemInf~\citep{deminf} estimates state--action mutual information.
CUPID~\citep{cupid} uses policy rollouts to estimate demonstration influence on closed-loop performance; QoQ~\citep{qoq} uses influence estimates based on validation loss.
DataMIL~\citep{datamil} models how training-data selection affects a target policy metric.
Our CUPID implementation provides a policy-dependent comparison.

\textbf{Demonstration quality and evaluation.}
Prior work studies data quality and composition~\citep{robomimic,belkhale}.
Studies of injected defects distinguish defect detection from downstream curation utility and identify episode-length confounding~\citep{bedi1,bedi2,bedi3}.
We examine natural operator variation and connect score definitions to subset selection and policy performance.
Prior findings on policy selection and statistical uncertainty~\citep{robomimic,precipice} also motivate our evaluation-protocol comparison.

\section{CuraEval}\label{sec:setup}
\begin{figure}[t]
\centering
\includegraphics[width=\linewidth]{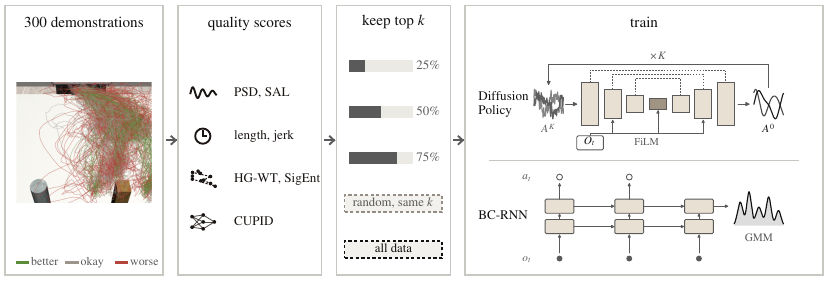}
\caption{\textbf{Overview of CuraEval.}
Candidate scores guide selection at 25/50/75\% retention from 300 mixed-skill demonstrations.
The top-$k$ illustration represents ranking-based selection and diversity adaptations also use greedy selection.
Policies trained on selected subsets are compared with size-matched random subsets and all data under fixed budgets within each task--policy setting.
Diffusion Policy is the main tested policy, with selected BC-RNN comparisons.}
\label{fig:overview}
\vspace{-5mm}
\end{figure}

\subsection{Evaluation Framework}
CuraEval examines what demonstration scores measure, how the selected data affect policy learning, and how evaluation protocols affect reported gains. 
Figure~\ref{fig:overview} presents an overview of CuraEval, from demonstration scoring and subset selection to policy training, with size-matched random subsets and full-data training as baselines.
For a dataset $\mathcal{D}$ of $N$ demonstrations, a selector retains $m$ demonstrations at retention level $r=m/N$.
We compare the resulting policy with policies trained on equally sized random subsets, simple-heuristic selections, and all available data.
Training hyperparameters and gradient-step budgets remain fixed within each task--policy setting.

\subsection{Scores and Selection Procedures}
PSD~\citep{psd} sums squared magnitudes of the unnormalized discrete Fourier transform (DFT) of raw end-effector (EE) positions over all frequency bins.
SAL follows RINSE's spectral arc length specification~\citep{rinse}.
Heuristics include episode length, total spectral power of EE velocity, and mean squared jerk.
Offline diagnostics additionally include path inefficiency: path length divided by start-to-end displacement.
For PSD and these heuristics, lower values are preferred, while SAL and diversity contributions prefer higher values.

We denote our whole-trajectory adaptation of Geometric Entropy~\citep{hg} by \textbf{\hg{}}, where WT stands for whole trajectory.
It resamples each complete EE trajectory to 50 points by arc length, translates its endpoint to the origin, and normalizes by start-to-end displacement.
It omits the original transit segmentation and target-orientation alignment.
Our \textbf{\sig{}} adaptation uses FAKTUAL's signature-entropy objective~\citep{faktual} with EE positions, time normalized to $[0,1]$, and a depth-3 signature kernel normalized to unit diagonal.

Both adaptations use LOO ranking or forward-greedy selection.
For a dataset-level objective $H$, the LOO contribution is
\begin{equation}
s_i^{\mathrm{LOO}}=H(\mathcal{D})-H(\mathcal{D}\setminus\{\tau_i\}).
\label{eq:loo}
\end{equation}
LOO computes these contributions once on the full dataset and retains the $m$ largest.
Greedy selection repeatedly adds the remaining demonstration that maximizes the resulting subset's objective.
We apply LOO ranking and forward-greedy selection to this adapted geometric-entropy measure.
\sig{} greedy maximizes entropy alone, omitting the full FAKTUAL procedure's determinant-based stage.

We implement CUPID~\citep{cupid} for robomimic DP, using TRAK-projected gradients in 4,000 dimensions and \texttt{sum-of-sum} aggregation: summed influence on successful rollouts minus that on failed rollouts.
Averaging demonstration ranks from three independently trained scoring policies gives one ranking and one fixed subset per retention level.
Each scoring policy uses 100 rollouts and approximately 12 A40-GPU-hours, giving 36 GPU-hours for the ensemble before downstream training.
The resulting subsets remain fixed across downstream training seeds.
Our diversity adaptations and CUPID implementation are evaluated under shared controls. 

\section{Experiments}\label{sec:experiments}
\subsection{Experimental Settings}\label{sec:experimental_settings}
We use successful multi-human demonstrations from robomimic~\citep{robomimic}.
\textit{can} and \textit{square} each contain 300 demonstrations from six operators across three skill levels, while \textit{transport} contains 300 bimanual demonstrations across six operator-skill pairings.
Our main experiments use state-based Diffusion Policy (DP)~\citep{dp} on \textit{can}/\textit{square}, supplemented by a 25\%-retention comparison on \textit{transport} and selected \textit{square} conditions with BC-RNN-GMM~\citep{robomimic}.
BC-RNN tests PSD, SAL, and length at 50/75\% retention, and the two diversity LOO selectors at 25\%, each with three training seeds and the corresponding controls.
The additional \hg{}-greedy comparison uses three seeds at 25\% retention on each of \textit{can} and \textit{square}.
Retention levels of 25/50/75\% correspond to 75/150/225 demonstrations and matching is by demonstration count, not transition count.
Within each task--policy setting, hyperparameters and training budgets are fixed: 50k steps for DP on \textit{can}/\textit{square}, 150k on \textit{transport}, and 200k for BC-RNN.
Each main-matrix random subset remains fixed across training seeds.
At 25\% retention on \textit{square}, two additional independent random subsets each receive three training runs.
The main matrix and these additional random-subset runs comprise 267 policy-training runs and 44,100 evaluation episodes.
These totals exclude six supplementary \hg{}-greedy runs and the scoring-policy rollouts.

We evaluate final checkpoints with 200 rollouts on \textit{square}/\textit{transport} and 100 on \textit{can}, using evaluator seed 424242 across conditions.
Evaluation horizons are 400, 500, and 700 steps for \textit{can}, \textit{square}, and \textit{transport}, respectively.
Each condition has three or six training seeds.
Training seed is the analysis unit for two-sided Welch's $t$-tests~\citep{welch1947} and pointwise 95\% confidence intervals, conditional on the selected subsets and evaluation settings.
Holm correction~\citep{holm1979} applies separately to comparisons against random and full data, including random controls in the latter.
Family sizes are 25/28 for \textit{square}-DP, 18/21 for \textit{can}-DP, and 8/11 for \textit{square}-BC-RNN.
Main significance statements use adjusted $\ph$, while supplementary comparisons are descriptive.
The \textit{transport} panel reports unadjusted exploratory tests because it is outside the three main comparison families.
The PSD--length and \textit{transport} comparisons were planned before their runs; the Holm families were specified retrospectively.
The study is exploratory: evaluation was revised and additional training runs were added after initial results.

\subsection{What Do Demonstration Scores Measure?}\label{sec:measure}\label{sec:psd}
For the evaluated PSD score~\citep{psd}, let $x_i(t)$ be an end-effector position sample, $T_i$ the number of samples, and $X_{i,d}(k)$ the unnormalized discrete Fourier transform of coordinate $d$.
Parseval's identity gives
\begin{equation}
W_i=\sum_{k,d}|X_{i,d}(k)|^2
=T_i\sum_t\|x_i(t)\|^2
=T_i^2\,\overline{\|x_i\|^2}.
\label{eq:parseval}
\end{equation}
This score is unchanged by reordering the same positions, so it cannot distinguish smooth from irregular temporal orderings.
The result concerns this full-spectrum position score, not spectral metrics generally.

The mean squared position magnitude varies little within each task (coefficient of variation 3.4--4.5\%), giving $W_i\approx cT_i^2$ for a task-dependent $c$.
Spearman correlations~\citep{spearman1904} with episode length are 0.998/0.999/0.993 on \textit{can}/\textit{square}/\textit{transport}.
PSD and length select the same 150 demonstrations on \textit{square} at 50\% retention.
Across eight downstream comparisons including six task--retention combinations with DP and two retention levels with BC-RNN, their mean success differences are at most 0.030 in magnitude, with all pointwise intervals including zero. This does not establish equivalence.
The identity is exact, while the correspondence with length depends on the position distribution.
Centering positions or using velocities changes the measured energy, but summing over all frequencies still discards its distribution across frequency bins.

Other rankings differ: velocity-PSD and jerk correlate 0.858 and $-0.240$ with the length ranking on \textit{square}.
The two diversity LOO rankings also agree inconsistently across \textit{can}, \textit{square}, and \textit{transport} ($\rho=0.57/0.44/0.04$).
Neither the heuristics nor the diversity objectives uniformly favor the same demonstrations.

\subsection{Downstream Policy Performance}\label{sec:selection}
\begin{table}[!t]
\centering
\caption{\textbf{Diffusion Policy on \textit{square}.}
Final checkpoints use 200 evaluation rollouts per training seed.
$n$: training seeds. Success: mean success probability.
Diff. is selection minus the indicated baseline, computed from unrounded means.
Random uses the original fixed size-matched subset. All data uses all 300 demonstrations.
Bold $p_H<0.05$ marks a significant gain or loss after Holm correction of two-sided Welch tests (25 comparisons against Random while 28 against All data).
Dashes denote inapplicable comparisons.}

\label{tab:main_square}
\normalsize
\setlength{\tabcolsep}{4pt}
\renewcommand{\arraystretch}{1.08}
\begin{tabular*}{\linewidth}{@{\extracolsep{\fill}}lrrrrrr@{}}
\toprule
Selection & $n$ & Success & \multicolumn{2}{c}{vs. Random} & \multicolumn{2}{c}{vs. All data} \\
\cmidrule(lr){4-5}\cmidrule(l){6-7}
& & & Diff. & $p_H$ & Diff. & $p_H$ \\
\midrule
All data (300) & 6 & 0.797 & --- & --- & --- & --- \\
\midrule
\multicolumn{7}{@{}l}{\textbf{Retain 75 of 300 demonstrations (25\%)}}\\
PSD & 3 & 0.660 & $+.035$ & $1.000$ & $-.137$ & $\boldsymbol{.011}$ \\
SAL & 3 & 0.633 & $+.008$ & $1.000$ & $-.163$ & $.155$ \\
Length & 3 & 0.630 & $+.005$ & $1.000$ & $-.167$ & $\boldsymbol{<.001}$ \\
CUPID & 3 & 0.628 & $+.003$ & $1.000$ & $-.168$ & $\boldsymbol{.020}$ \\
HG-WT (LOO) & 3 & 0.417 & $-.208$ & $\boldsymbol{.026}$ & $-.380$ & $\boldsymbol{<.001}$ \\
SigEnt (LOO) & 3 & 0.470 & $-.155$ & $.057$ & $-.327$ & $\boldsymbol{.003}$ \\
SigEnt (greedy) & 3 & 0.580 & $-.045$ & $1.000$ & $-.217$ & $\boldsymbol{<.001}$ \\
Random & 3 & 0.625 & --- & --- & $-.172$ & $\boldsymbol{.011}$ \\
\midrule
\multicolumn{7}{@{}l}{\textbf{Retain 150 of 300 demonstrations (50\%)}}\\
PSD & 6 & 0.760 & $+.053$ & $.071$ & $-.037$ & $.584$ \\
SAL & 6 & 0.758 & $+.052$ & $.086$ & $-.038$ & $.584$ \\
Length & 6 & 0.765 & $+.058$ & $\boldsymbol{.026}$ & $-.032$ & $.753$ \\
Velocity-PSD & 3 & 0.728 & $+.022$ & $1.000$ & $-.068$ & $.769$ \\
Jerk & 6 & 0.776 & $+.069$ & $\boldsymbol{.022}$ & $-.021$ & $1.000$ \\
CUPID & 3 & 0.772 & $+.065$ & $.121$ & $-.025$ & $1.000$ \\
HG-WT (LOO) & 3 & 0.685 & $-.022$ & $1.000$ & $-.112$ & $\boldsymbol{.006}$ \\
SigEnt (LOO) & 3 & 0.732 & $+.025$ & $1.000$ & $-.065$ & $1.000$ \\
SigEnt (greedy) & 3 & 0.723 & $+.017$ & $1.000$ & $-.073$ & $.584$ \\
Random & 6 & 0.707 & --- & --- & $-.090$ & $\boldsymbol{.004}$ \\
\midrule
\multicolumn{7}{@{}l}{\textbf{Retain 225 of 300 demonstrations (75\%)}}\\
PSD & 6 & 0.796 & $+.033$ & $.323$ & $-.001$ & $1.000$ \\
SAL & 6 & 0.807 & $+.044$ & $.320$ & $+.010$ & $1.000$ \\
Length & 6 & 0.804 & $+.042$ & $.125$ & $+.008$ & $1.000$ \\
Velocity-PSD & 6 & 0.824 & $+.062$ & $\boldsymbol{.037}$ & $+.028$ & $1.000$ \\
Jerk & 3 & 0.790 & $+.028$ & $1.000$ & $-.007$ & $1.000$ \\
CUPID & 3 & 0.820 & $+.058$ & $.171$ & $+.023$ & $1.000$ \\
HG-WT (LOO) & 3 & 0.770 & $+.008$ & $1.000$ & $-.027$ & $1.000$ \\
SigEnt (LOO) & 3 & 0.743 & $-.019$ & $1.000$ & $-.053$ & $.140$ \\
SigEnt (greedy) & 6 & 0.741 & $-.022$ & $.504$ & $-.056$ & $.065$ \\
Random & 6 & 0.763 & --- & --- & $-.034$ & $.584$ \\
\bottomrule
\end{tabular*}
\vspace{-5mm}
\end{table}

Several simple heuristics outperform random selection on \textit{square} (Table~\ref{tab:main_square} presents the detailed quantitative results).
At 50\% retention, jerk and length selection increase DP success from 70.7\% to 77.6\% ($\ph=0.022$) and 76.5\% ($\ph=0.026$), respectively.
At 75\%, velocity-PSD increases success from 76.3\% to 82.4\% ($\ph=0.037$).
Positive estimates for PSD, SAL, and CUPID do not survive the same correction.
Length selection at 50\% nevertheless remains below the 79.7\% full-data mean.
No curated condition significantly outperforms full-data training in any of the three main task--policy settings.
Nevertheless, velocity-PSD and CUPID at 75\% reach 82.4\% and 82.0\% on \textit{square}; their success differences from full data have pointwise 95\% intervals $[-0.0091,0.0641]$ and $[-0.0147,0.0613]$.
Useful improvements therefore remain compatible with the observations.

\begin{table}[!t]
\centering
\caption{\textbf{Diffusion Policy on \textit{can} and \textit{transport}.}
Final checkpoints use 100 and 200 evaluation rollouts per training seed, respectively.
$n$: training seeds. Success: mean success probability.
Diff. is selection minus the indicated baseline, computed from unrounded means.
Random uses the original fixed size-matched subset. All data uses all 300 demonstrations.
Bold $p_H<0.05$ marks a significant gain or loss after Holm correction of two-sided Welch tests on \textit{can} (18 comparisons against Random while 21 against All data).
The \textit{transport} panel reports unadjusted exploratory tests.}

\label{tab:main_can}
\normalsize
\setlength{\tabcolsep}{4pt}
\renewcommand{\arraystretch}{1.08}
\begin{tabular*}{\linewidth}{@{\extracolsep{\fill}}lrrrrrr@{}}
\toprule
Selection & $n$ & Success & \multicolumn{2}{c}{vs. Random} & \multicolumn{2}{c}{vs. All data} \\
\cmidrule(lr){4-5}\cmidrule(l){6-7}
& & & Diff. & $p_H$ & Diff. & $p_H$ \\
\midrule
All data (300) & 6 & 0.967 & --- & --- & --- & --- \\
\midrule
\multicolumn{7}{@{}l}{\textbf{Retain 75 of 300 demonstrations (25\%)}}\\
PSD & 3 & 0.980 & $+.033$ & $.873$ & $+.013$ & $.904$ \\
SAL & 3 & 0.967 & $+.020$ & $1.000$ & $+.000$ & $1.000$ \\
Length & 3 & 0.977 & $+.030$ & $.873$ & $+.010$ & $1.000$ \\
HG-WT (LOO) & 3 & 0.740 & $-.207$ & $\boldsymbol{.011}$ & $-.227$ & $\boldsymbol{.029}$ \\
SigEnt (LOO) & 3 & 0.850 & $-.097$ & $.100$ & $-.117$ & $\boldsymbol{<.001}$ \\
SigEnt (greedy) & 3 & 0.853 & $-.093$ & $.540$ & $-.113$ & $.492$ \\
Random & 3 & 0.947 & --- & --- & $-.020$ & $1.000$ \\
\midrule
\multicolumn{7}{@{}l}{\textbf{Retain 150 of 300 demonstrations (50\%)}}\\
PSD & 6 & 0.983 & $+.013$ & $1.000$ & $+.017$ & $.684$ \\
SAL & 6 & 0.985 & $+.015$ & $1.000$ & $+.018$ & $.684$ \\
Length & 6 & 0.993 & $+.023$ & $.682$ & $+.027$ & $.099$ \\
HG-WT (LOO) & 3 & 0.920 & $-.050$ & $.051$ & $-.047$ & $\boldsymbol{.029}$ \\
SigEnt (LOO) & 3 & 0.947 & $-.023$ & $.682$ & $-.020$ & $.371$ \\
SigEnt (greedy) & 3 & 0.973 & $+.003$ & $1.000$ & $+.007$ & $1.000$ \\
Random & 6 & 0.970 & --- & --- & $+.003$ & $1.000$ \\
\midrule
\multicolumn{7}{@{}l}{\textbf{Retain 225 of 300 demonstrations (75\%)}}\\
PSD & 6 & 0.980 & $+.005$ & $1.000$ & $+.013$ & $1.000$ \\
SAL & 6 & 0.993 & $+.018$ & $.341$ & $+.027$ & $.099$ \\
Length & 6 & 0.992 & $+.017$ & $.456$ & $+.025$ & $.122$ \\
HG-WT (LOO) & 3 & 0.947 & $-.028$ & $.682$ & $-.020$ & $1.000$ \\
SigEnt (LOO) & 3 & 0.953 & $-.022$ & $.682$ & $-.013$ & $1.000$ \\
SigEnt (greedy) & 3 & 0.977 & $+.002$ & $1.000$ & $+.010$ & $1.000$ \\
Random & 6 & 0.975 & --- & --- & $+.008$ & $1.000$ \\
\bottomrule
\end{tabular*}
\medskip
\textbf{\textit{transport}: exploratory comparisons}\par\smallskip
\begin{tabular*}{\linewidth}{@{\extracolsep{\fill}}lrrrrrr@{}}
\toprule
Selection & $n$ & Success & \multicolumn{2}{c}{vs. Random} & \multicolumn{2}{c}{vs. All data} \\
\cmidrule(lr){4-5}\cmidrule(l){6-7}
& & & Diff. & Raw $p$ & Diff. & Raw $p$ \\
\midrule
All data (300) & 3 & 0.335 & --- & --- & --- & --- \\
Random (75) & 3 & 0.165 & --- & --- & $-.170$ & 0.004 \\
HG-WT (LOO, 75) & 3 & 0.098 & $-.067$ & 0.069 & $-.237$ & 0.004 \\
\bottomrule
\end{tabular*}
\vspace{-5mm}
\end{table}

On \textit{can}, full-data success is already 96.7\% (Table~\ref{tab:main_can} presents the detailed quantitative results).
Length selection at 50/75\% and SAL at 75\% achieve 99.2--99.3\%, but their improvements survive neither the random-control nor the full-data correction.
The smallest adjusted $p$ for a positive full-data difference is 0.099.
With BC-RNN on \textit{square}, the highest observed 75\%-retention mean is 67.5\% for SAL, compared with 67.2\% for full data, with substantial training-seed uncertainty.
Performance retention would require a prespecified acceptable loss and a non-inferiority analysis.

The largest losses occur at 25\% retention with adapted diversity measures.
\hg{}-LOO lowers DP success from 62.5\% to 41.7\% on \textit{square} ($\ph=0.026$) and from 94.7\% to 74.0\% on \textit{can} ($\ph=0.011$).
With BC-RNN on \textit{square}, \hg{}-LOO and \sig{}-LOO lower success from 28.3\% to 17.5\% ($\ph=0.003$) and 17.7\% ($\ph=0.002$).
The \textit{transport} estimate is also negative (9.8\% versus 16.5\%), but its interval includes zero. This small, under-converged comparison remains inconclusive.

The three independently sampled random subsets on \textit{square} achieve 62.5\%, 60.7\%, and 59.8\% mean success.
Both \hg{}-LOO (41.7\%) and \sig{}-LOO (47.0\%) remain below every random-subset mean.
Each draw is held fixed across its three training runs, so the experiment comprises three independent subset replicates, not nine.
The seed rosters differ across subsets; this descriptive check does not isolate subset variance or change the original Holm families.

\subsection{What Do the Selectors Retain?}\label{sec:div}
\begin{figure}[t]
\centering
\includegraphics[width=\linewidth]{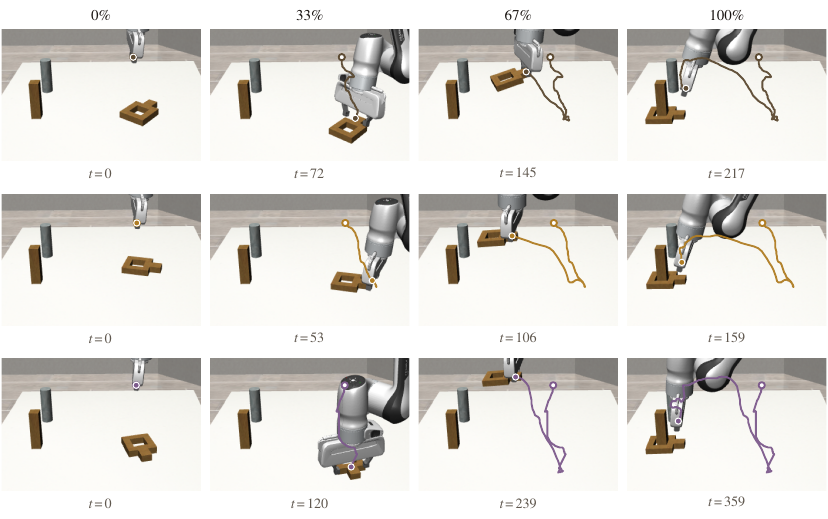}
\caption{\textbf{Successful demonstrations retained by different selectors.}
Top to bottom: Random subset 0 (\texttt{demo\_191}, 218 steps), Length (\texttt{demo\_43}, 160 steps), and \hg{}-LOO (\texttt{demo\_69}, 360 steps).
Each is the median-length demonstration among its 75 selected on \textit{square} at 25\% retention, with ties broken by numerical ID.
Columns show 0/33/67/100\% of each episode's time index and actual timesteps appear below.
A shared camera and crop show the accumulated end-effector path. Open/filled markers denote start/current position.
Colors identify selectors: the first two examples are higher-skill, the third lower-skill.
All are successful human demonstrations, not policy rollouts.}
\label{fig:qualitative}
\vspace{-5mm}
\end{figure}

Figure~\ref{fig:qualitative} shows retained demonstrations.
On \textit{square}, \hg{}-LOO retains 51 lower-skill and only two higher-skill demonstrations out of 75.
Their mean episode length is 403 samples, compared with 269 for the full dataset.
Length selection instead retains three lower-skill and 60 higher-skill demonstrations, with a mean length of 159 samples.
However, operator composition alone does not explain the outcomes: \sig{}-LOO and entropy-greedy retain the same counts of 16 higher-skill, 24 middle-skill, and 35 lower-skill demonstrations, yet achieve 47.0\% and 58.0\% success.
Skill-matched or length-matched controls would be needed to isolate these factors from other differences between selected trajectories.

Selection procedures also matter within the geometric-entropy adaptation.
Forward-greedy selection achieves 49.0\% on \textit{square} and 72.0\% on \textit{can}, compared with 41.7\% and 74.0\% for LOO at 25\% retention.
Both procedures remain below the original random controls (62.5\% and 94.7\%).
Changing the selector affects success, but neither procedure makes these adapted objectives consistently beneficial at this budget.
These additional greedy comparisons are descriptive and do not extend the main Holm families.

CUPID's merged ranking correlates $-0.14$ with length, while its three scorer rankings have pairwise correlations of 0.16--0.31.
Both scoring policies and rollout samples vary.
At 25\% retention, pairwise Jaccard overlap between the three selected sets is 0.22--0.24, compared with an approximate random-set reference of 0.14.
These observations make scoring stability relevant alongside the additional computational cost.

\subsection{Sensitivity to Evaluation Protocols}\label{sec:protocol}
Our initial protocol reported the maximum success estimate across training checkpoints, using 50 rollouts each.
The updated protocol evaluates the final checkpoint with 200 rollouts and a common evaluator seed.
We compare the same three training seeds in each policy class: DP seeds 1/2/3 and BC-RNN seeds 0/1/2.
The additional DP seeds used in the main matrix are excluded from this comparison.

On \textit{square}, length selection at 75\% retention initially increases DP success from 76.7\% for full-data training to 86.7\%.
Under the updated protocol, the corresponding means are 80.5\% for full data and 79.5\% for length selection.
BC-RNN shows the same reversal: its original full-data and selected means are 76.0\% and 82.7\%, versus 67.2\% and 66.2\% under the updated protocol.
The updated differences have pointwise 95\% Welch intervals $[-0.0515,0.0315]$ and $[-0.1866,0.1666]$, respectively: the estimates reverse sign without establishing significant harm.
Checkpoint reporting, rollout count, and evaluation seed change jointly; their individual effects are not isolated.

Reporting procedures differ in published work: FAKTUAL takes a maximum over 20 robomimic checkpoints with 50 rollouts each~\citep{faktual}, while CUPID averages the last ten simulation checkpoints with 50 rollouts each~\citep{cupid}.
Our results support explicit reporting of these choices and independent testing after checkpoint selection, or a prespecified endpoint or averaging rule.

\section{Discussion and Limitations}\label{sec:limits}
Curation evaluations should distinguish selecting better data than random, improving over full-data training, and retaining performance with fewer demonstrations.
The last goal requires a practical tolerance: a non-significant loss alone is insufficient.
Exact representations, selection rules, and evaluation procedures are necessary to interpret each comparison.
Fixed gradient-step budgets measure performance under the same number of updates, rather than training time to a target success rate.
Smaller subsets are revisited more often, and matching demonstration counts does not match transition counts.
Equal transition or runtime budgets would address complementary uses of curation.

Our evidence is limited to small state-based datasets and the stated adaptations and policy configurations.
\textit{can} is near ceiling and \textit{transport} remains under-converged.
Our full-data DP success on \textit{square} is 79.7\%, versus 87.0\% under the PSD paper's different implementation and data split~\citep{psd}. 
This discrepancy remains unresolved.
Only \textit{square}-DP at 25\% retention tests multiple random subsets.
Common evaluator seeds do not measure uncertainty across independently drawn evaluation sets.
Full FAKTUAL, phase-aligned Geometric Entropy, visual policies, VLA fine-tuning, and real robots remain as our future work.

\section{Conclusion}
CuraEval combines score analysis with controlled selection experiments.
The evaluated PSD score closely tracks episode length, simple heuristics can outperform random selection, and aggressive selection with our diversity adaptations can substantially reduce success.
The \textit{square} losses persist across three random subsets.
No curated condition has a Holm-adjusted full-data advantage, although useful gains remain compatible with the intervals.
Joint protocol changes reverse the estimated benefit of length selection.
These findings motivate reporting the selected data, selection rule, and evaluation protocol together.

\end{document}